\documentclass[letterpaper, 10 pt, conference]{ieeeconf}  

\IEEEoverridecommandlockouts                              

\usepackage{cite}
\usepackage{amsmath,amssymb,amsfonts}
\usepackage{bm}
\usepackage{booktabs}
\usepackage[inkscapearea=drawing, inkscapepath=figures/generated, inkscapelatex=false]{svg}
\usepackage[on, extractpath=out/figures]{svg-extract}
\usepackage[caption=false, font=footnotesize]{subfig}
\usepackage{algorithm}
\usepackage{algorithmic}
\usepackage{graphicx}
\usepackage{textcomp}
\usepackage{xcolor}
\usepackage{tikz}
\let\labelindent\relax 
\usepackage{enumitem}
\usepackage[hidelinks]{hyperref}
\usepackage{todonotes}
\usepackage[symbol,hang,flushmargin]{footmisc}
\usepackage[normalem]{ulem}
\usepackage[nolist]{acronym}
\usepackage{listings}
\usepackage{float}
\usepackage{comment}
\usepackage{subfig}
\acrodef{spi}[SPI]{Shadow Program Inversion}
\acrodef{dof}[DoF]{degree of freedom}
\acrodefplural{dof}[DoF]{degrees of freedom}
\acrodef{rl}[RL]{Reinforcement Learning}
\acrodef{tcp}[TCP]{tool center point}
\acrodef{dmp}[DMP]{Dynamic Movement Primitive}
\acrodef{prodmp}[ProDMP]{Probabilistic Dynamic Movement Primitive}
\acrodef{iql}[IQL]{Implicit Q-Learning}
\acrodef{awr}[AWR]{Advantage Weighted Regression}
\acrodef{bc}[BC]{Behaviour Cloning}
\acrodef{dcg}[DCG]{differentiable computation graph}
\acrodef{vit}[ViT]{Vision Transformer}
\acrodef{xai}[XAI]{explainable artificial intelligence}
\acrodef{dp}[$\partial$P]{differentiable programming}
\acrodef{ann}[ANN]{artificial neural network}
\acrodef{mutt}[MuTT]{Multimodal Trajectory Transformer}

\definecolor{lbcolor}{rgb}{0.95,0.95,0.95}  
\title{\LARGE \bf MuTT: A Multimodal Trajectory Transformer for Robot Skills}

\author{Claudius Kienle$^{1}$, Benjamin Alt$^{1}$, Onur Celik$^{2}$, Philipp Becker$^{2}$, Darko Katic$^{1}$\\ Rainer Jäkel$^{1}$ and Gerhard Neumann$^{2}$
\thanks{This  work  was  supported  by  the  German  Federal  Ministry  of  Education and Research (grant 01MJ22003B).}
\thanks{$^{1}$ArtiMinds Robotics, Karlsruhe, Germany}%
\thanks{$^{2}$Autonomous Learning Robots, KIT, Germany}%
}

\newcommand\copyrighttext{%
  \footnotesize \textcopyright 2024 IEEE. Personal use of this material is permitted.
  Permission from IEEE must be obtained for all other uses, in any current or future 
  media, including reprinting/republishing this material for advertising or promotional 
  purposes, creating new collective works, for resale or redistribution to servers or 
  lists, or reuse of any copyrighted component of this work in other works.}
\newcommand\copyrightnotice{%
\begin{tikzpicture}[remember picture,overlay]
\node[anchor=south,yshift=10pt] at (current page.south) {\fbox{\parbox{\dimexpr\textwidth-\fboxsep-\fboxrule\relax}{\copyrighttext}}};
\end{tikzpicture}%
}

\begin{document}
\bstctlcite{IEEEexample:BSTcontrol}

\maketitle
\copyrightnotice
\thispagestyle{empty}
\pagestyle{empty}

\begin{abstract}

High-level robot skills represent an increasingly popular paradigm in robot programming. However, configuring the skills' parameters for a specific task remains a manual and time-consuming endeavor. Existing approaches for learning or optimizing these parameters often require numerous real-world executions or do not work in dynamic environments. To address these challenges, we propose \ac{mutt}, a novel encoder-decoder transformer architecture designed to predict environment-aware executions of robot skills by integrating vision, trajectory, and robot skill parameters. Notably, we pioneer the fusion of vision and trajectory, introducing a novel trajectory projection. Furthermore, we illustrate \ac{mutt}'s efficacy as a predictor when combined with a model-based robot skill optimizer. This approach facilitates the optimization of robot skill parameters for the current environment, without the need for real-world executions during optimization. Designed for compatibility with any representation of robot skills, \ac{mutt} demonstrates its versatility across three comprehensive experiments, showcasing superior performance across two different skill representations. 

\end{abstract}

\section{Introduction}
\label{sec:introduction}

The use of robots in industry depends to a large extent on the difficulty of programming the robot for the task to be solved. One programming paradigm that has become increasingly popular is programming with high-level robot skills \cite{johannsmeier_framework_2019,marvel_automated_2009,thomas_new_2013,pedersen_robot_2016}. 
While there exist plentiful skill representations, like \acp{dmp} \cite{ijspeert_dynamical_2013,li_prodmp_2023,schaal_dynamic_2006} or classical force-controlled skills \cite{bruyninckx_specification_1996}, all of them have in common that they have a set of parameters (e.g. goal pose, robot velocity) to configure them for the task at hand.
Choosing the correct parameters can be a time-consuming and complex process, further complicated by the fact that the optimal parameterization of a skill is highly sensitive to both the robot and its environment. Prior work on learning~\cite{hussein_imitation_2018} or optimizing task parameters \cite{alt_robot_2021} is limited by the need for hours of real-world robot executions or fails to generalize to environment changes.

In light of these challenges, we propose \ac{mutt}, an innovative encoder-decoder transformer for environment-aware predictions of robot skill executions. To the best of our knowledge, \ac{mutt} stands out as the first multimodal transformer that fuses trajectory and vision modalities, enabling trajectory predictions based on visual reasoning. 
A trajectory is a defined path in Joint or Cartesian space the robot follows over a specified period of time. Every point in this path may encompass additional information, such as the forces and torques experienced, the task success, or the expected reward for \ac{rl} agents.
Moreover, \ac{mutt} can be used as a predictor of real-world executions in robot skill optimizers \cite{alt_robot_2021}, enabling the optimization of robot skills for the current environment without requiring real-world executions during optimization.

With this, our contribution is threefold:
\begin{enumerate}
    \item \ac{mutt}: A transformer that can easily be finetuned to predict environment-aware trajectories of robot skills. This includes a novel projection for trajectories that retains pose and force information at high temporal resolution, enabling precise prediction of high-frequency features such as sharp peaks.
    \item Integration of \ac{mutt} as a predictive model into an optimizer for skill-based robot programs.
    \item Evaluation of the prediction and optimization capabilities of \ac{mutt} on two real-world industrial tasks and one ManiSkill2 environment \cite{gu_maniskill2_2023}, using two different skill frameworks \cite{li_prodmp_2023,bruyninckx_specification_1996}.
\end{enumerate}

\begin{figure}
    \centering
    \includesvg[width=\linewidth]{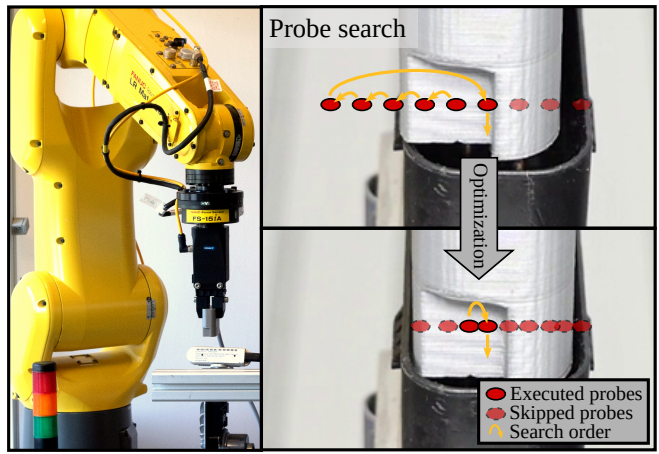}
    \caption{\ac{mutt} is used in the \ac{spi} parameter optimizer \cite{alt_robot_2021} to refine the initial search pattern (red dots, top). The optimization yields an improved search pattern (red dots, bottom), reducing the required probes from six to two to successfully locate the socket. This significantly decreases the cycle time of the task. As an environment-aware model of robot skills, \ac{mutt} enables the optimization of skill parameters for the current environment, alleviating the need for real-world executions during optimization.}
    \label{fig:mutt-spike-search}
\end{figure}

\section{Related Work}
\label{sec:related_work}

\subsection{Multimodal Transformer}
Transformers were applied to many multimodal application domains in the past, typically combining the modalities vision and text \cite{le_vision_2021,wu_masked_2022,khare_mmbert_2021,kim_vilt_2021}. 

Recent approaches are delving into the integration of text and audio \cite{ao_speecht5_2022}, as well as audio and vision \cite{shi_learning_2022,li_visualbert_2019}. Nevertheless, these approaches consistently operate under the assumption of modal synchronization. For instance, in~\cite{shi_learning_2022}, video processing involves aligned audio tracks and video frames. This exploitation of modality synchronization enables the consolidation of synchronized segments from both streams into a singular token. 
Gemini 1.5 \cite{reid_gemini_2024} is a notable exception as being one of the first to combine audio and vision without requiring modal synchronization. However, as the architecture of Gemini 1.5 is not publicly available, there is no information about its implementation.

In the robotics domain, \cite{brohan_rt-2_2023,ghosh_octo_nodate} have advanced the training of a multimodal transformer architecture for closed-loop robot control. 
Similarly, \cite{driess_palm-e_2023} integrate language instructions and images, although their model focuses on predicting text-based instructions rather than directly predicting raw actions.

While numerous approaches exist that combine vision, audio, or text \cite{radford_learning_2021,khare_mmbert_2021,ao_speecht5_2022,kim_vilt_2021,brohan_rt-2_2023,ghosh_octo_nodate,driess_palm-e_2023,hsu_hubert_2021,li_visualbert_2019,brohan_rt-1_2023,reid_gemini_2024,shi_learning_2022}, there currently lacks a multimodal transformer that effectively merges trajectory and vision modalities. 
Despite the potential similarity between the audio and trajectory modalities due to both being continuous-valued time-series, trajectories require a more precise resolution which hinders the use of discretization, often done for the projection of audio data \cite{ao_speecht5_2022}. Additionally, the trajectory length is crucial, which for audio is often not considered~\cite{hsu_hubert_2021},~\cite{ao_speecht5_2022}.

\subsection{Forward Dynamics Models}

Previous approaches \cite{lim_real2sim2real_2022,sukhija_gradient-based_2023,du_learning_2023} develop a forward dynamics model to predict the state of the robot or environment given a robot action. The learned model is then utilized to determine the action that yields the optimal state, using either sample-based methods or gradient-based optimization. For instance, \cite{du_learning_2023} propose a forward model that, given a text-based description of a task, predicts a video depicting the robot performing the task. From the video, they derive the robot's action sequence. 
However, most forward dynamics models are tailored to specific tasks, such as predicting the cable tip in \cite{lim_real2sim2real_2022}. In contrast, \ac{mutt} receives and predicts trajectories, allowing for its application across a wide range of tasks and robot skill representations. 
Furthermore, \ac{mutt} is the first forward model to incorporate an environment image, enabling it to predict the robot state conditioned on the robot's current environment.

\subsection{Robot Skill Optimization}

Alt et al. \cite{alt_robot_2021} propose a model-based optimizer for skill-based robot programs that learns a differentiable model of a sequence of robot skills. This ``shadow program'' predicts the expected robot execution given the skills' parameters, enabling the optimization of robot skill parameters via gradient descent without additional executions on the real robot. Alternative approaches optimize skill parameters with Reinforcement Learning \cite{zhang_online_2022}, Bayesian Optimization \cite{johannsmeier_framework_2019}, or Evolutionary Algorithms \cite{marvel_automated_2009} by repeatedly executing the program with varying parameterizations on a real robot or in a simulated environment for evaluation.

Most of the proposals for robot skill optimization \cite{marvel_automated_2009,johannsmeier_framework_2019} require executing candidate parameterizations on the real robot to assess them for the current environment.  This significantly slows down the optimization and necessitates a real robot during optimization. Additionally, it poses the risk of potentially damaging the robot or its surroundings, as the chosen program parameterizations are solely determined by the optimizer, potentially leading to unforeseen consequences during execution. A simulated environment can be beneficial, but introduces the Sim2Real Gap \cite{hofer_sim2real_2021}, which complicates the transfer of learned skills to the real robot. While \cite{alt_robot_2021} avoids this by learning a surrogate model that predicts the robot execution, this predictor has no sense of the current environment the robot operates in, restricting its application to static environments.

Our contribution incorporates an environmental signal and enables the prediction of accurate robot trajectories for the current environment. With this, it is not tailored to one skill representation or one specific optimization algorithm, but can be integrated as an environment-aware predictor in gradient-based \cite{alt_robot_2021,alt_heuristic-free_2022}, gradient-free \cite{kennedy_particle_1995,shahriari_taking_2016,back_overview_1993} and model-based~\cite{berkenkamp_bayesian_2020,calandra_bayesian_2016} optimization strategies. This eliminates the need to perform real robot executions during optimization of the robot skill.

\section{\acl{mutt}}

\begin{figure*}
    \centering
    \includesvg[width=\textwidth]{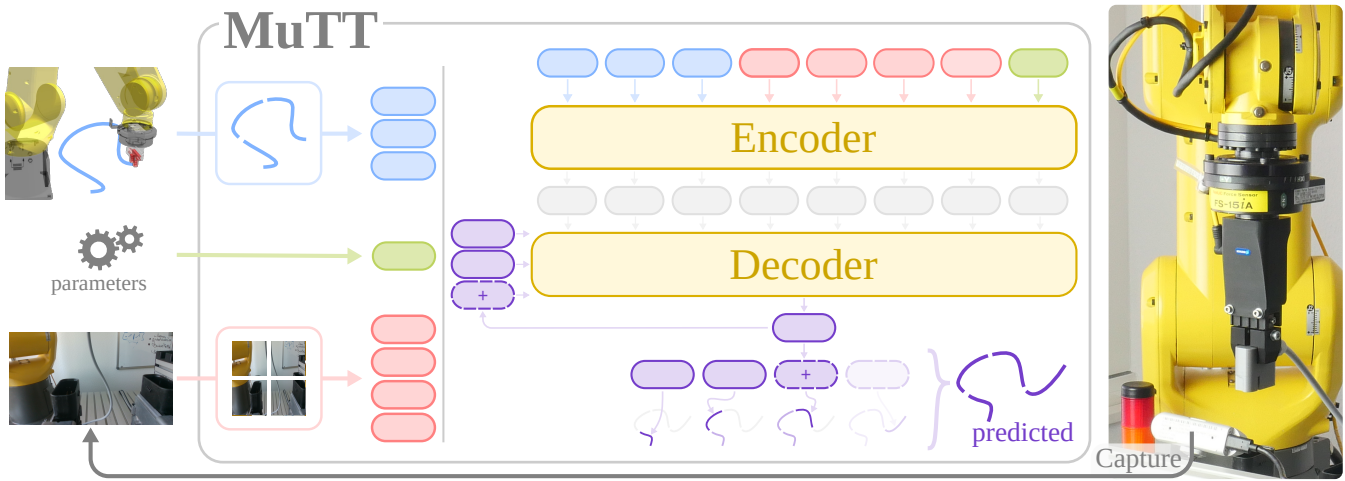}
    \caption{\acl{mutt} architecture: modality specific embedding of the simulated trajectory (blue), skill parameters (green) and environment image (red) into tokens, which are concatenated to one token sequence. All tokens are coded with modality specific positional and token-type encoding and passed through an encoder transformer. The decoder predicts the real-world trajectory (purple) in an autoregressive manner.}
    \label{fig:mutt-architecture}
\end{figure*}

\ac{mutt} is an encoder-decoder transformer \cite{vaswani_attention_2017} that predicts the execution of robot skills for the current environment. The encoder fuses the different modalities into one hidden representation, based on which the decoder predicts the environment-aware trajectory autoregressively. The model architecture is depicted in Figure \ref{fig:mutt-architecture}.

\subsection{Encoder}

The encoder of \ac{mutt} follows the \textit{single-stream} approach~\cite{bugliarello_multimodal_2021} and uses a minimal embedding pipeline to project the trajectories, environment images and skill parameters into a token sequence. As we are the first to consider the trajectory modality, we propose a novel trajectory embedding further detailed in \ref{sec:methods:traj-prediction}. An image of the environment is split into 16x16 patches and linearly projected into tokens as in~\cite{dosovitskiy_image_2021}. We interpolate over the position encoding for the image tokens to enable the processing of differently sized images \cite{dosovitskiy_image_2021}. The robot skill parameters are embedded by a single linear layer. All embedded tokens are coded with modality-specific positional encoding and their respective token type before they are passed through the encoder transformer to obtain a fused hidden representation. 

\subsection{Decoder}

The \ac{mutt} decoder predicts the trajectory segment-wise in an autoregressive manner, while feeding the already predicted trajectory segments back into the decoder alongside the encoder's hidden representation, as in \cite{ao_speecht5_2022}. We decide on a segment-wise prediction and not a prediction of individual points to utilize the hidden state effectively and reduce the inference duration and number of autoregressive iterations. A segment size of 20 points proved to be the best, as it did not reduce the prediction accuracy, but significantly reduced the memory requirements and inference time of \ac{mutt}. The decoder should predict the trajectory that matches the execution of the robot skill in a particular environment. Both the encoder and decoder have a hidden size of 768, the encoder consists of 12 attention layers, while the decoder consists of 6 attention layers. Every attention layer uses 12 attention heads.

We chose an encoder-decoder architecture to enable the prediction of variable-length trajectories and to support a variable number of image patches.
An encoder-only or decoder-only transformer would require padding input or output to a fixed length, artificially enlarging the data that must be processed. In Experiment \ref{sec:experiment-grasp-skill}, we compare the encoder-decoder transformer with an encoder-only transformer. The encoder-only transformer predicted trajectories whose length deviated significantly more from the execution than the predicted trajectories of the encoder-decoder transformer. Moreover, a fixed input or prediction size limits the ability to transfer a trained \ac{mutt} instance to a new task that would require processing longer trajectories. 

\subsection{Trajectory Projection}
\label{sec:methods:traj-prediction}

Trajectories are continuous-valued time-series with a fixed temporal sampling interval. In comparison to many other modalities, trajectories need a high prediction accuracy and also require predicting the exact length of the trajectory. We normalize trajectories by the dataset mean and standard deviation. Since the trajectories in a batch can be of different length, we pad them to the length of the longest in the batch by replicating the last point. Additionally, the batch is padded to a multiple of 20 points. Padding a trajectory involves annotating every point with a binary flag that indicates whether the point serves as padding or not. We split the trajectories along the time dimension into equal sized segments of 20 points. The trajectory embedding follows the minimal embedding pattern used in \cite{dosovitskiy_image_2021} and projects the segments with a single linear layer in a token sequence. In contrast to related approaches working with actions \cite{brohan_rt-2_2023,janner_offline_2021}, the trajectories are not discretized before projection, which would reduce their resolution. We also do not apply any smoothing or additional pre- or postprocessing to the trajectories. Furthermore, we set the attention mask value for a token to zero if the entire segment consists of padding points only and otherwise to one. The position of every token is encoded with absolute position encoding \cite{gehring_convolutional_2017}. Relative positional encoding or interpolation would violate the invariant that there is a fixed time interval between two successive data points in a trajectory.

\subsection{Pre-training}

\ac{mutt} must be trained to predict the real execution of a robot skill. Since collecting a real-world dataset of robot executions is costly, we aim to minimize the data required to train the model effectively. Consequently, we compared initializing the transformer encoder with weights from related models \cite{ao_speecht5_2022,kim_vilt_2021,dosovitskiy_image_2021,hsu_hubert_2021}. Nearly all initialization variants improved the performance of the finetuned \ac{mutt} significantly, except for \cite{hsu_hubert_2021}. The initialization with \ac{vit} \cite{dosovitskiy_image_2021} resulted in the best performance. We use the weights of SpeechT5 \cite{ao_speecht5_2022} for the decoder transformer due the similarity between the audio and trajectory modality, which stems from their shared characteristic as continuous-valued time series.

\section{Experiments}

\begin{figure*}
    \centering
    \includesvg[width=\textwidth]{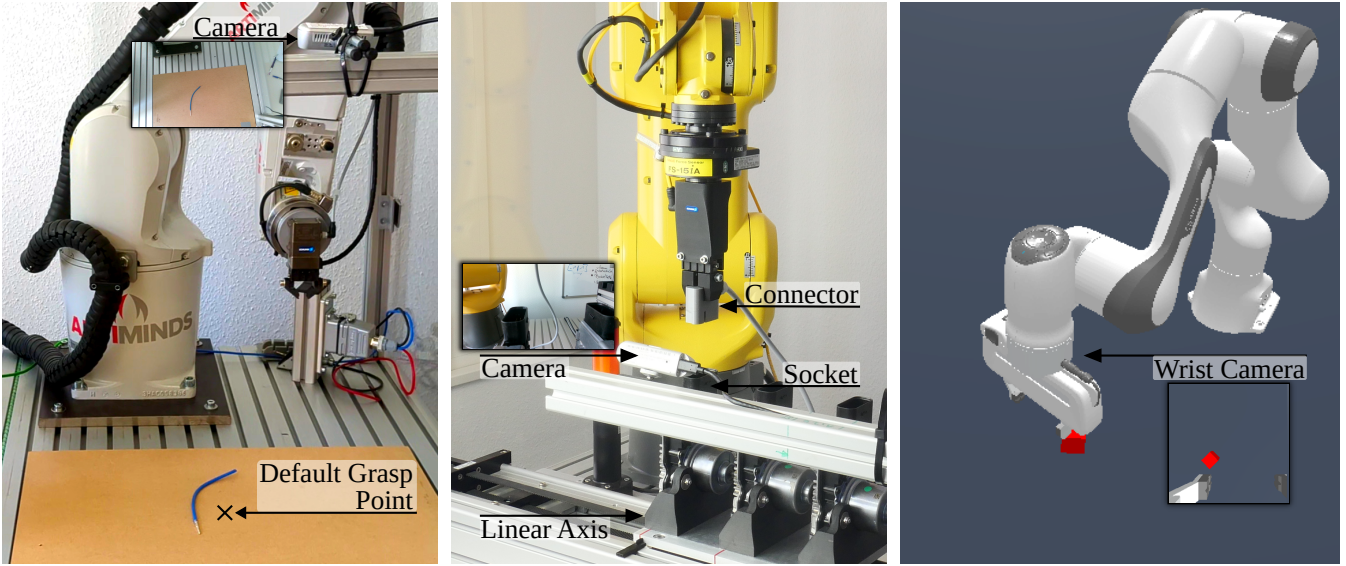}
    \caption{Evaluation scenarios: Real-world grasping of deformable cables (Experiment \ref{sec:experiment-grasp-skill}, left), real-world force-controlled plug insertion under uncertainty (Experiment \ref{sec:experiment-plug-skill}, middle), and simulated grasping in the ManiSkill2 benchmark (Experiment \ref{sec:experiment-maniskill}, right).}
    \label{fig:experiment-overview}
\end{figure*}

We demonstrate the capability of \ac{mutt} to accurately predict the real-world execution of robot skills by evaluating it in three different manipulation scenarios, using two different robot skill representations. This showcases \ac{mutt}'s design to seamlessly work with various robot skill representations, without necessitating any modifications to the model architecture or training algorithm.
An overview of the experiments is shown in Figure \ref{fig:experiment-overview}. In all experiments, we embed \ac{mutt} as an environment-aware predictor of real robot-skill executions in existing frameworks \cite{alt_robot_2021,gu_maniskill2_2023} to optimize the robot skills for the current environment.

\subsection{Model-Based Optimization of Industrial Robot Skills}

We use \ac{mutt} to optimize industrial robot skills \cite{bruyninckx_specification_1996} for grasping deformable cables and force-controlled plug insertion. In both experiments, \ac{mutt} receives a simulated trajectory of the skill, the skill's parameters, and an unprocessed RGB environment image from a RealSense D435. Based on these inputs, the model must predict the real-world trajectory resulting from the execution of the skill. The trained \ac{mutt} instances are then used in the robot program parameter optimization framework \ac{spi} \cite{alt_robot_2021} to optimize the program for the current environment.

\subsubsection{Grasping of Deformable Cables}
\label{sec:experiment-grasp-skill}
\begin{figure}
    \centering
    \includesvg[width=\linewidth]{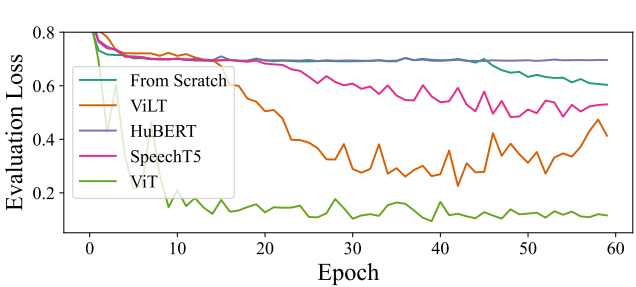}
    \caption{Comparison of training \ac{mutt} on the dataset of Experiment \ref{sec:experiment-grasp-skill} with different initial weights from related applications \cite{ao_speecht5_2022,dosovitskiy_image_2021,kim_vilt_2021,hsu_hubert_2021}. Initialization with ViT \cite{dosovitskiy_image_2021} resulted in the best evaluation performance.}
    \label{fig:pretrain-comparison}
\end{figure}

This experiment is designed to clearly demonstrate that \ac{mutt} can be used as forward model in existing optimization frameworks, in this case for the optimization of robot skill parameters with \ac{spi} \cite{alt_robot_2021}.
\begin{figure}
    \centering
    \includesvg[width=\linewidth]{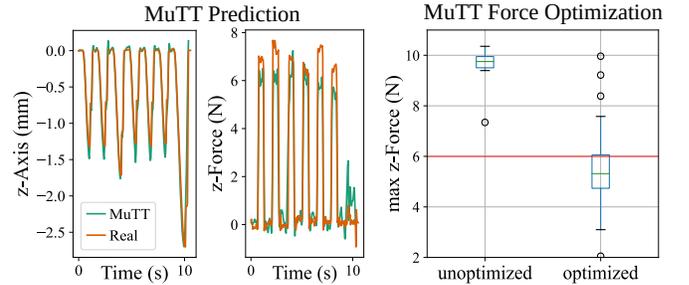}
    \caption{Experiment \ref{sec:experiment-plug-skill}: End-effector pose (along Z axis, left) and force (along Z axis, middle) prediction of \ac{mutt} for a probe skill. \ac{mutt} predicts forces accurately, enabling the optimization with \ac{spi} \cite{alt_robot_2021} of the robot skill to adhere to the user-defined force limit of 6 N, while the unoptimized skill greatly exceeds that limit (right).}
    \label{fig:experiment-plug-prediction}
\end{figure}
In the experiment, the robot tries to grasp an industrial cable from a table with a grasp skill parameterized by a grasp pose. The cable's position varies by up to 5 centimeters in each direction. \ac{mutt} is trained on 5,000 randomly selected grasp executions. We compare different initialization strategies for \ac{mutt} \cite{dosovitskiy_image_2021,kim_vilt_2021,hsu_hubert_2021,ao_speecht5_2022}, with ViT \cite{dosovitskiy_image_2021} yielding the best evaluation performance, as seen in Figure \ref{fig:pretrain-comparison}. \ac{mutt} predicts the execution of the grasp skill, including whether the cable was grasped successfully. Subsequently, we utilized \ac{mutt} to predict the real robot executions within the parameter optimizer \ac{spi}~\cite{alt_robot_2021}, optimizing the grasp pose for successful cable grasping. While the robot grasped the cable in 2 of 100 evaluation runs with initial parameterizations sampled from the region of feasible grasp poses, parameter optimization with \ac{mutt} and \ac{spi} led to grasping the cable in 67 of the same 100 evaluation runs. 


This demonstrates \ac{mutt}'s capability to integrate the environment image and program parameters to accurately predict the likelihood of grasping in the current environment. With this, the grasp pose can be optimized until \ac{mutt} predicts a successful grasp of the cable.

\subsubsection{Force-Controlled Plug Insertion}
\label{sec:experiment-plug-skill}
The second experiment is representative of many current applications of robots in industry and shows that \ac{mutt} is able to predict complex skill executions. In this experiment, the predictions not only include the end-effector pose, but also the forces and torques that occur at the  end-effector during execution. 
Specifically, we study the insertion of an industrial connector into the matching socket with a force-controlled probe-search skill. In real-world industrial applications, the exact positioning of parts such as the socket is often subject to process noise. We simulate random deviations of the positioning of the socket by up to 1 cm with a linear axis. The robot searches for the socket along the linear axis. In this experiment, \ac{mutt} predicts the end-effector motion during search, which deviates from the planned motion due to environment interaction, This includes the unknown position of the socket, on which the robot's search motion depends. Additionally, \ac{mutt} predicts the end-effector forces and torques during search, and a success token indicating whether the robot successfully plugged the connector into the socket. \ac{mutt} must adjust its prediction based on the socket's position depicted in the environment image and the search pattern given by the skill parameters and simulated trajectory. 

We train \ac{mutt} on 5,000 randomly parameterized robot skill executions. The trained \ac{mutt} instance accurately predicts the search motion of the end-effector with an average deviation to the real execution by 0.3 mm and an average force deviation of 0.5 N. The success of the search was predicted correctly with an F1 score of 0.99. Figure~\ref{fig:experiment-plug-prediction} depicts an exemplary prediction of \ac{mutt} alongside the real execution of that same robot skill.

We show that the accurate prediction of \ac{mutt} can be used in \ac{spi} to optimize the search skill, including the search pattern for a fast and successful search, as well as the contact forces during the search. In force-controlled skills, reaction delays consistently lead to the robot exceeding pre-set force limits. Consequently, optimizing the search skill to adhere to a user-defined force-threshold safeguards hardware (such as the grasped plug) from damage by exceeding the force limit.

\begin{table}
    \centering
    \caption{Experiment \ref{sec:experiment-plug-skill}: Results}
    \label{tab:experiments:plug:optim-eval}
    \begin{tabular}{l c c}
        \toprule
                                              & Unoptimized & Optimized \\
        \midrule
        Avg number of probes                  & 14          & 3         \\
        Avg search duration (s)               & 25.3       & 7.5      \\
        Success probability                     & 0.55        & 0.96      \\
        Avg exceeded forces by (N)            & 3.5        & 0.7      \\
        \bottomrule
    \end{tabular}
\end{table}

Table \ref{tab:experiments:plug:optim-eval} compares the execution of the initial robot skill and the optimized robot skill on the real robot. \ac{mutt} predicts the real execution accurately for the current deviation of the socket. This enabled \ac{spi} to optimize the robot skill parameters resulting in a 70~\% faster search, doubling the success probability while adhering to the force limit 78 \% more accurately. The maximal forces experienced during search with unoptimized and optimized parameters can be seen in Figure \ref{fig:experiment-plug-prediction}. While using the unoptimized parameters resulted in exceeding the user-defined force threshold by about 4 N, the optimized programs kept to the force limit.

\subsection{Model-Based Optimization of \ac{prodmp} Skills}

\begin{figure}
    \centering
    \includesvg[width=.90\linewidth]{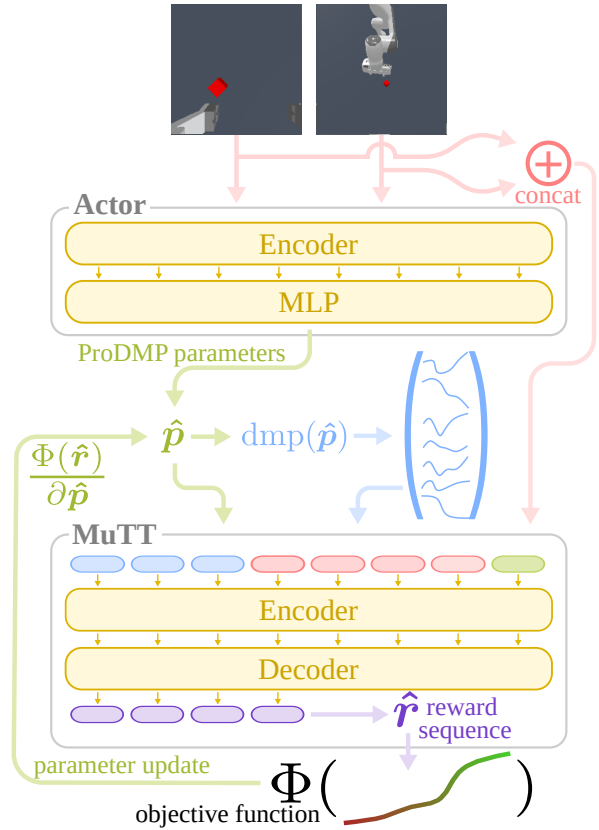}
    \caption{Architecture of Experiment \ref{sec:experiment-maniskill}. The \ac{bc} actor predicts \ac{prodmp} parameters $\bm{\hat{p}}$ that define a trajectory to pick up the red cube shown in the environment images. \ac{mutt} predicts the expected reward $\bm{\hat{r}}$ the agent would receive for every action in this trajectory. $\Phi$ computes the mean squared distance to the maximal reward the actor can receive for an action. Via gradient descent, the \ac{prodmp} parameters $\bm{\hat{p}}$ are optimized to minimize $\Phi$ and consequently maximize the predicted reward $\bm{\hat{r}}$, resulting in improved \ac{prodmp} parameters~$\bm{\hat{p}}$.}
    \label{fig:experiment:maniskill:architecture}
\end{figure}

In a second series of experiments, we apply \ac{mutt} to \acp{prodmp}, demonstrating that its architecture is not limited to one specific robot skill representation.
\acp{prodmp} provide a representation capable of generating smooth trajectories from any initial state while capturing higher-order motion statistics.
This experiment assesses \ac{mutt}'s prediction capabilities on the \textit{PickCube-v0} environment of the ManiSkill2 Benchmark \cite{gu_maniskill2_2023}. We focus solely on picking up the cube from the surface without evaluating whether the cube was lifted to the correct position.
Here, \ac{mutt}'s predictions are used to optimize the trajectory suggested by an episodic \ac{rl} agent, resulting in an increase of task success compared to the raw agent. For this, we learn a robot skill with an episodic \acf{bc} actor that predicts parameters $\bm{\hat{p}}$ for 7 \acp{prodmp} \cite{li_prodmp_2023}, 6 of which define the end-effector pose (position and rotation) and one the gripper configuration. \ac{mutt} receives the \ac{prodmp} parameters, the simulated trajectory generated by the \acp{prodmp}, and concatenated images of two cameras as input. In contrast to the first two experiments, the real-world trajectory predicted by \ac{mutt} also contains the reward $\bm{\hat{r}}$ the agent would receive for every action in the given \ac{prodmp} trajectory. This underscores \ac{mutt}'s versatility, as the predicted trajectory is not limited to the robots motion, but can contain any sequence of data associated with the execution of the robot skill. Via gradient descent, the parameters $\bm{\hat{p}}$ are optimized to maximize the reward $\bm{\hat{r}}$ predicted by \ac{mutt}. The architecture is illustrated in Figure~\ref{fig:experiment:maniskill:architecture}. 

\subsubsection{ManiSkill2 Benchmark}
\label{sec:experiment-maniskill}

We train a \ac{bc} actor \cite{karamcheti_language-driven_2023} on a dataset $\mathcal{D}_{demo}$ consisting of 1,000 optimal demonstrations and use the trained actor to generate an additional dataset $\mathcal{D}_{im}$ consisting of 30,000 imperfect executions. Subsequently, \ac{mutt} is trained on $\mathcal{D}_{im}$ to predict the expected reward for every action. Finally, we employ the trained \ac{mutt} instance to optimize the \acp{prodmp} predicted by the \ac{bc} actor. To achieve this, we compute a loss $\Phi(\bm{\hat{r}})$ based on the reward $\bm{\hat{r}}$ 
\begin{equation*}
    \Phi(\bm{\hat{r}}) := \frac{1}{|\bm{\hat{r}}|} \sum_{i=1}^{|\bm{\hat{r}}|} (r_{max} - \hat{r}_i) ^ 2
\end{equation*}
and update the \ac{prodmp} parameters $\bm{\hat{p}}$ via gradient descent. $r_{max}$ is the maximal reward the actor can receive for one action, in this experiment set to $1$. This aims to maximize the reward $\bm{\hat{r}}$ predicted by \ac{mutt}. 

\begin{figure}
    \centering
    \includesvg[width=.78\linewidth]{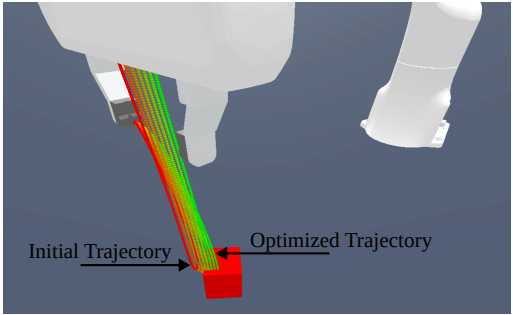}
    \caption{Robot skill optimization in Experiment \ref{sec:experiment-maniskill}. Evolution of the trajectory generated from the \ac{prodmp} parameters throughout the optimization process. The robot would not have grasped the cube with the initial robot skill parameters (red). Over the course of the optimization, the trajectory is gradually aligned with the cube's position. Ultimately, with the optimized \ac{prodmp} parameters, the robot successfully grasps the cube~(green).}
    \label{fig:experiment:maniskill:optimization}
\end{figure}

Table \ref{tab:experiment:maniskill} compares the \ac{bc} actor with and without optimization by \ac{mutt} to the offline episodic \ac{rl} algorithm \ac{awr} \cite{kurkova_episodic_2008} trained on $\mathcal{D}_{im}$ and to the state-action based offline \ac{rl} algorithm \ac{iql} \cite{kostrikov_offline_2021}, trained on 66,000 state-action transitions generated from the $\mathcal{D}_{im}$ episodic dataset. All algorithms are evaluated on the same 100 environments they have not seen during training. 

Optimizing the \acp{prodmp} predicted by the \ac{bc} actor with \ac{mutt} increases the number of successful task executions by 6 \%. Figure \ref{fig:experiment:maniskill:optimization} displays how the optimization improves the end-effector trajectory leading to successful grasping of the cube, while the robot does not grasp the cube with the trajectory predicted by the \ac{bc} actor. \ac{mutt} notably excels the \ac{awr} algorithm that struggles to learn the task based on the same dataset $\mathcal{D}_{im}$ \ac{mutt} was trained on, likely due to only 30 \% of the samples in this dataset successfully solving the task. \ac{mutt} also dominates \ac{iql}, which was trained on a dataset more than twice as large. While it is difficult to compare episodic and state-action based algorithms, it demonstrates that \ac{mutt} outperforms state-action based algorithms that were trained for a comparable number of steps on a dataset of comparable size. Additionally, state-action based algorithms can adapt their prediction for every new state online, while \ac{mutt} predicts the entire trajectory given the initial state, without any online adaptation during execution.

\begin{table}[]
    \centering
    \caption{Experiment \ref{sec:experiment-maniskill}: Results}
    \label{tab:experiment:maniskill}
    \begin{tabular}{l c c c}
        \toprule
          & Time steps & Dataset size & Success prob. \\
          \midrule
         \ac{bc} \cite{karamcheti_language-driven_2023} & 60k & 1k & 0.33 \\
         \ac{bc} + \ac{mutt} (Ours) & 7.3M & 30k & \textbf{0.39}  \\
         \ac{awr} \cite{kurkova_episodic_2008} & 15M & 30k & 0.02 \\
         \ac{iql} \cite{kostrikov_offline_2021} & 9.3M & 66k & 0.35 \\ 
         \bottomrule
    \end{tabular}
\end{table}

\section{Conclusion and Outlook}
\label{sec:conclusion}
We introduce \ac{mutt}, a \acl{mutt} that accurately predicts robot skill executions aligned with the robot's current environment by integrating vision, trajectory, and robot skill parameters. \ac{mutt} is representation-agnostic with respect to the robot skill and can be applied to near-arbitrary skill representations. To that end, we developed a novel trajectory projection that retains important properties such as the trajectories' temporal resolution and length.

\ac{mutt}'s architecture allows for efficient training with relatively small datasets of random skill executions, making it a promising foundation model with quick adaptation to specific robot skills. Furthermore, \ac{mutt}'s compatibility with any robot skill optimizer enables the optimization of robot program parameters tailored to the current environment.

While \ac{mutt} as predictor for model-based robot skill optimization offers significant advantages over traditional optimization methods, such as not requiring real-world executions during optimization and precise trajectory prediction in dynamic environments, some challenges remain. The prediction accuracy of \ac{mutt} on out-of-distribution samples should be further analyzed. First tests showed that \ac{mutt}'s capability to generalize well to such samples is limited, indicating room for improvement through future research. Additionally, the speed of parameter optimization relies heavily on the inference speed of \ac{mutt}. While we engineered the model for fast and efficient inference, the optimization of robot skills currently takes about 20 to 40 seconds. Optimizing the model architecture to decrease inference duration and consequently enhance optimization speed is open for future research.

\bibliographystyle{IEEEtran}  
\bibliography{bibliography_config,bibliography}

\end{document}